\documentclass[10pt,a4paper]{article}
\usepackage{epsfig}
\usepackage{amsmath,amsthm,amsfonts,amssymb}
\usepackage[T1]{fontenc}
\usepackage[latin1]{inputenc}
\def\v{\mathbf{v}}
\def\z{\mathbf{z}}
\def\vx{\mathbf{v}_x}
\def\vy{\mathbf{v}_y}

\def\z{\mathbf{z}}

\def\0{\mathbf{0}}
\def\sigt{\sigma_T^2}
\def\Cxy{\tilde{\Sigma}_{xy}}
\def\Cxx{\tilde{\Sigma}_{xx}}
\def\Cyy{\tilde{\Sigma}_{yy}}

\title{Dependency detection with similarity constraints}

\author{ Leo Lahti\(^{1,2}\), Samuel Myllykangas\(^3\), Sakari
Knuutila\(^2\) and Samuel Kaski\(^1\).\\ 1. Helsinki University of
Technology, Department of Information and\\ Computer Science, PO Box
5400, FI-02015 TKK, Finland\\ 2. University of Helsinki and Helsinki
University Central Hospital,\\ Haartman Institute and HUSLAB,
Department of Pathology, Helsinki, Finland.\\ 3. Stanford University
School of Medicine, Department of Medicine,\\ Division of Oncology,
and Stanford Genome Technology Center,\\ Stanford University,
Stanford, USA.}

\date{}

\begin{document}

\maketitle

\subsubsection*{Preprint of Lahti {\em et al.} 2009 in T{\"u}lay
Adali, Jocelyn Chanussot, Christian Jutten, and Jan Larsen, editors,
Proceedings of the 2009 IEEE International Workshop on Machine
Learning for Signal Processing XIX, pages 89--94. IEEE, Piscataway,
NJ, USA, 2009. Derivations presented in more detail in
http://lib.tkk.fi/Diss/2010/isbn9789526033686\\ Implementation of the
method available at\\
http://bioconductor.org/packages/devel/bioc/html/pint.html}

\begin{abstract}

  Unsupervised two-view learning, or detection of dependencies between
  two paired data sets, is typically done by some variant of canonical
  correlation analysis (CCA). CCA searches for a linear projection for
  each view, such that the correlations between the projections are
  maximized. The solution is invariant to any linear transformation of
  either or both of the views; for tasks with small sample size such
  flexibility implies overfitting, which is even worse for more
  flexible nonparametric or kernel-based dependency discovery
  methods. We develop variants which reduce the degrees of freedom by
  assuming constraints on similarity of the projections in the two
  views. A particular example is provided by a cancer gene discovery
  application where chromosomal distance affects the dependencies
  between gene copy number and activity levels. Similarity
  constraints are shown to improve detection performance of known
  cancer genes.

\end{abstract}

\section{Introduction}
\label{sec:intro}

We develop methods for the task of detecting statistical dependencies
between multiple sources of co-occurring data. The sources are assumed
to share relevant common information, and additionally contain
independent but unknown type of noise. The task is to discover the
relevant information; both to detect and analyse or interpret it.

This is a particular type of a data fusion task, shared by
\emph{multi-view learning}. In multi-view learning each source is
interpreted as a different view to the same items, and the task is to
enhance classification performance by combining the views. Our task
can be interpreted as unsupervised multi-view learning.

The traditional statistical way of finding dependencies between data
sources is canonical correlation analysis, CCA, which generalizes
correlation to multidimensional sources, retaining some of the nice
interpretability of correlation coefficients. While the basic
correlation coefficient assumes paired scalar values, canonical
correlations assume paired vectorial values. The vectors are projected
to scalar components before computing the correlations, using linear
projections that maximize the correlations. For multidimensional data
there will be many correlation coefficients; the second components are
constrained to be uncorrelated with the first, and so on.

CCA is known to have two nice properties: the result is invariant to
linear transformations of the data spaces, and the solution for any
fixed number of components maximizes mutual information between linear
projections for Gaussian data. These insights can be interpreted as
motivations for generalizing using nonparametric methods
\cite{Fisher00,Klami05icassp} and kernel CCA \cite{Fyfe00,Bach05}.

The flexibility of CCA can cause overfitting problems that are
specifically harmful with small sample sizes that abound in biomedical
studies, for instance. When the views are high-dimensional, the
completely unconstrained linear projections involve high degrees of
freedom; several ways to regularize the CCA solution have been
suggested to overcome some of the associated problems \cite{Bie03,
  Sun08, Vinod76}. We introduce a complementary approach that is based
on bringing in prior knowledge to constrain the model family.

Assuming the dimensions of the different views are not completely
unrelated but instead are formed of related pairs, it makes sense to
search for more constrained projections. In our application, the views
are different measurements made on the same locations of the genome,
and the dimensions correspond to these particular
locations. Constraining the projections to be the same or at least
similar in the different views will additionally enhance
interpretability of the results, given that relationships between the
same components in the two views are natural.

Correlation-based CCA has been shown to correspond to the maximum likelihood
solution of a simple generative model \cite{Bach05}, where the two
views are assumed to stem from a shared Gaussian latent variable and
normally distributed data-set-specific noise. This has opened up the
road to probabilistic and Bayesian formulations
\cite{Klami06mlsp,Klami07icml} which make it possible to deal
rigorously with uncertainty in small sample sizes and to include prior
knowledge as Bayesian priors.

We suggest also a probabilistic version for constrained dependency
search that provides a robust alternative for direct maximization of
correlations.  While the probabilistic version is slower to compute,
it is the recommended choice when prior information of the types of
dependency is available, or sample size is small.

The methods will be applied in a very promising application setup for
knowledge discovery with dependency detection. The task is to find
potential cancer genes by studying the relationship between changes
caused by cancer in gene expression and gene copy numbers, that is,
amplifications or deletions caused by mutations in cancer samples.
Copy number changes are a key mechanism for cancer, and combination of
copy number information with gene expression measurements can reveal
functional effects of the mutations; gene expression data is
informative of gene activity.  The rationale goes as follows:
Mutations having no functional effect will not cause cancer, and
cancer-related gene expression changes may be side effects. Gene
expression changes caused by mutations would be strong candidates for
cancer mechanisms, and they contribute to the dependencies between the
two data sources.  While causation can be difficult to grasp, study of
the dependencies can provide an efficient proxy for such effects.

\section{Canonical correlations with similarity constraints}
\label{sec:method}

\subsection{Correlation-based approach}

Correlation-based CCA searches for a maximally correlated linear
projection of the original data sets with paired samples \(X\) and
\(Y\).  It maximizes the correlation between the projections,
\(cor(X\vx,Y\vy)\), with respect to arbitrary projection vectors
\(\vx,\vy\). However, this flexibility easily leads to overfitting as
demonstrated by the case study in Section~\ref{sec:application}.

In many applications prior information of the potential relationships
between the features of the investigated data sets is
available. Constraining the projections accordingly can potentially
reduce overfitting and help to focus on specific types of dependencies
between the two data sets. A particular example of such a model is
provided by our cancer gene discovery application, where gene copy
number changes are systematically correlated with the gene expression
measurements from the same genes. 

The relationship between the projections can be para\-metr\-ized with
a transformation matrix \(T\) such that \(\vy = T \vx\). Maximization
of the correlations between the projections leads to the following
optimization problem:
\begin{equation}\label{eq:simcca}
\arg \max_{\v, T} = \frac{\v^T \Cxy T\v}{ \sqrt{\v^T \Cxx \v} \sqrt{(T\v)^T \Cyy T\v}},
\end{equation}
where the observed covariances of the two data sets are denoted by the
\(\tilde{\Sigma}\). Constraints on \(T\) can be used to guide
the dependency search. We refer to this model as
Similarity-constrained CCA ({\it SimCCA}). Suitable constraints depend
on the particular applications; the solutions can be made to prefer
particular types of dependencies in a soft manner with an appropriate
penalty term on \(T\). 

While we consider only one-dimensional projections in the case study,
multidimensional projection matrices are also possible. The optimal
projection vectors can be sought iteratively as in ordinary
CCA. Direct optimization of the correlations provides a simple and
computationally efficient way to detect dependencies between data
sources but it lacks an explicit model to deal with the uncertainty in
the data and model parameters.

\subsection{Probabilistic approach}

An explicit model-based approach for the dependency exploration task
is provided by the probabilistic modeling fra\-me\-work. We derive a
probabilistic approach which should be more robust to small sample
sizes.  The correlation-based CCA has a direct connection to the
maximum likelihood (ML) solution of the  generative model
\cite{Bach05,Archambeau06}:
\begin{equation}\label{eq:genmodel}
   \begin{array}{cl}
	X \sim N(W_x \z, \Psi_x)\\ 
	Y \sim N(W_y \z, \Psi_y),
  \end{array} 
\end{equation}
assuming normally distributed \(\z\), and data-set-specific
covariances \(\Psi_x, \Psi_y\). The dependency between the data sets
is captured by the shared latent variable \(\z\), and \(W_x, W_y\)
characterize the relationship between the data sets. The covariances
\(\Psi_x, \Psi_y\) characterize data set-specific effects. Note that
while optimal projections \(\v\) in the correlation-based CCA
(Eq.~\ref{eq:simcca}) operate on the observed data, the parameters
of interest, \(W_x, W_y\), in probabilistic CCA mediate
transformations of the latent variable \(\z\).

The solutions of the probabilistic CCA can be constrai\-ned
analogously to the correlation-based approach in Eq~(\ref{eq:simcca}),
by extending the formulation to include appropriate prior terms.  The
joint likelihood of the model is given by
\begin{flalign}\label{eq:probsimcca0}
P (X, & Y, W,\Psi) \\
\sim &  P(X,Y|W_x,W_y,\Psi)P(W_y|W_x)P(W_x)P(\Psi) \\
 = & \int P(X,Y|W_x,W_y,\Psi,\z)& \\
 & P(W_y|W_x)P(W_x)P(\Psi)P(\z) d\z.
\end{flalign} 
Here \(\Psi\) denotes the block-diagonal matrix of \(\Psi_x\) and
\(\Psi_y\). While incorporation of prior information of the data
set-specific effects through the \(W_x\) and \(\Psi\) provides
promising lines for further work, we focus on the shared latent
variables as a probabilistic alternative to the correlation-based
SimCCA. The relation between the transformation matrices for the shared latent variable is encoded by the prior term \(P(W_y|W_x)\) and
can be parametrized with a transformation matrix \(T\) such that \(W_y
= T W_x\). Assuming invertible \(W_x^T W_x\), we have \(T = W_y (W_x^T
W_x)^{-1} W_x^T\).

By setting a prior on \(T\) it is possible to emphasize certain types
of dependencies. With unconstrained \(T\) the solution reduces to
ordinary probabilistic CCA. In the other extreme \(T\) is an identity
matrix, \(T = I\), and the two shared components, derived from $x$ and
$y$ respectively, would be identical. The formulation would also allow
tuning of \(T\) between these two extremes.

We consider the following simple prior for \(T\): \(P(T) = N_{+}(
\parallel (T - I) \parallel | 0, \sigt) = N_{+}( \parallel W_y (W_x^T
W_x)^{-1} W_x^T) - I \parallel | 0, \sigt)\). This can be plugged into
\(P(W_y|W_x)\) in Eq.~(\ref{eq:probsimcca0}). We have used Frobenius
norm, and \(N_{+}\) refers to truncated normal distribution for
positive input values.

The \(\sigt\) can tune the deviation of \(T\) from the identity
matrix; a strict version of probabilistic SimCCA (pSimCCA) is obtained
with \(\sigt \rightarrow 0\), while \(\sigt \rightarrow \infty\)
yields ordinary probabilistic CCA (pCCA). With uninformative priors
\(P(W), P(\Psi) \sim 1\) and normally distributed shared latent
variable \(\z \sim N(0,I)\), the model has the negative
log-likelihood
\begin{equation}\label{eq:probsimcca}
  -logP(X,Y,W,\Psi) \sim log|\Sigma| + tr \Sigma^{-1} \tilde{\Sigma} + \frac{\parallel T - I  \parallel}{\hat{\sigt}}.
\end{equation}
Here \(\Sigma = WW^T + \Psi \) contains the matrices \(W_x, W_y\) and data
set specific covariances \(\Psi_x, \Psi_y\). We have added the prior
for \(T\), which tunes the relationship between \(W_y\) and
\(W_x\). For other details, see \cite{Bach05,Bie03}.

\section{Analysis of functional copy number changes in gastric cancer}
\label{sec:application}

A promising biomedical application highlights the potential practical
value of our approach. Constraints on the potential dependencies
between gene expression and copy number are shown to improve the
detection of known cancer genes. The advantages of constrained and
probabilistic versions become particularly salient when the
dimensionality increases and ordinary correlation-based CCA seriously
overfits to the data.

\subsection{Background and motivation}

Copy number changes in chromosomal regions with tu\-mor-suppressor or
other cancer-associated genes have important contribution to cancer
development and progression.  Chromosomal gains and losses are likely
to be positively correlated with the expression levels of the affected
genes; copy number gain is likely to increase the expression of some
of the associated genes whereas deletion will block gene expression.
Identification of cancer-associated regions with functional copy
number changes has potential diagnostic, prognostic and clinical
impact for cancer studies.

Canonical correlations provide a principled framework for detecting
the shared variation in gene expression and copy number data.
Systematic copy number changes in a particular chromosomal region are
captured by multiple copy number probes, and this is also visible in
the expression levels of the genes within the affected region. The
dependent signals can be subtle, however, as gene expression and copy
number data are affected by high levels of unrelated biological and
measurement variation, and the sample sizes are typically small.

Both correlation-based and probabilistic SimCCA combine power over the
adjacent genes by capturing the strong\-est shared signal in gene
expression and copy number observations. They can also ignore
unrelated signal from poorly performing probes, or probes that measure
genes that are not functionally affected by the copy number
change. This provides tools to distinguish between so-called driver
mutations having functional effects from less active passenger
mutations, which is an important task in cancer studies. A further
advantage of the probabilistic formulation is that the shared latent
variable \(\z\) provides a robust measure of the amplification effects
in each patient.

\begin{figure}[htb]
 \centerline{\rotatebox{270}{\epsfig{figure=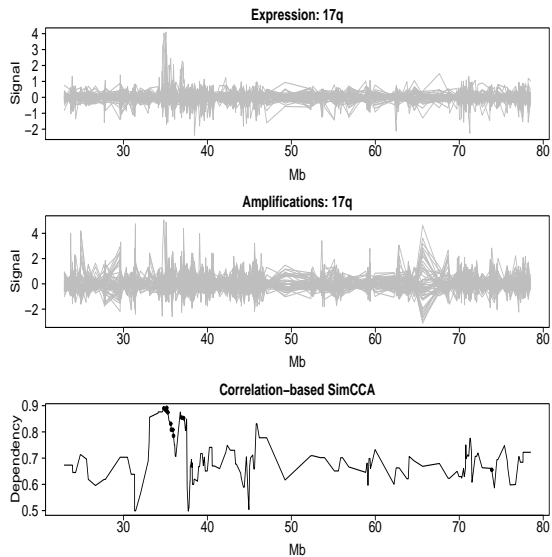,width=7.5cm,height=7.5cm}}}
 \caption{Gene expression, copy number signal, and the dependency
   score for a sliding window of 15 genes along the chromosome arm
   17q from the SimCCA method of Eq.~(\ref{eq:simcca}). Known
   gastric-cancer associated genes from an expert-curated list are
   marked with black dots.}
\label{fig:17q}
\end{figure}

\subsection{Implementation}

SimCCA is used to study the association between gene expression and
copy number in a gastric cancer data set with 41 patients and 10
controls \cite{Myllykangas08}. The gene expression and copy number
data sets were matched for the analysis such that the closest probe by
genomic location in gene expression data was selected for each copy
number probe, and probes with no match between gene expression and
copy number within 5000 bp interval were discarded. The preprocessed
data has gene expression and copy number measurements from 5596 genes
from \(\sim 700\) chromosomal regions (cytobands). To satisfy the
normality assumptions of our model, the data was \(log_2\)-transformed
and the mean of the signals for each probe was set to 0 before the
analysis.

Ordinary and constrained versions of canonical correlation analysis,
CCA/SimCCA, were applied to investigate the dependencies between gene
expression and copy numbers. The correlations were computed within a
specific chromosomal window around each gene. The observed
correlations provide a measure of dependency between gene copy number
and expression data for each window, or chromosomal region.

With unconstrained \(T\), the models defined by Eqs.~(\ref{eq:simcca}) and~(\ref{eq:probsimcca})  reduce to ordinary
correlation-based and probabilistic CCA, respectively.  We assume that
the constraints for \(T\) are provided prior to analysis, i.e. the prior parameter \(\sigma_T\) is fixed. Alternatively, \(\sigma_T\) could be optimized based on external criteria such as identification
of the known cancer genes in our application. Our empirical results
show, however, that already a simple prior for \(T\) without an
explicit optimization procedure can improve the detection of known
cancer genes.

We consider here the two extreme cases of the model where T is
(i) completely unconstrained (ordinary CCA; \(\sigma_T = \infty\)), and (ii) \(T=I\) (\(\sigma_T = 0\)). Point
estimates for the model parameters were estimated with EM algorithm in
the probabilistic version. Strength of the shared signal versus
marginal effects is measured with \(Tr(WW^T)/Tr(\Psi)\), where \(Tr\)
denotes matrix trace. This yields a dependency score between
copy number and expression data for the investigated chromosomal
neighborghood around each gene. High scores highlight regions where
the dependent signal between the two data sets is particularly high
relative to the data-set-specific variation.

In addition to the correlation-based and probabilistic SimCCA, we
tested a simplified probabilistic version with one-dimensional shared
component \(\z\) and isotropic covariances for the data-set-specific
effects: (\(\Psi_x = \sigma^2_x I; \Psi_y = \sigma^2_y I\)). This is a
special case of the full probabilistic model, and it reduces to
principal component analysis (PCA) for concatenated data
\((X,Y)\). We refer to this method as pSimPCA.  The
simplified model does not distinguish between the shared and marginal
effects as effectively as the full probabilistic CCA but it has fewer
model parameters. Low-dimensional latent models are also faster to
compute, and interpretation of the results is potentially more
straightforward.

\subsection{Validation}

\begin{figure}[htb]
\begin{minipage}[b]{1.0\linewidth}
  \centering
 \centerline{\rotatebox{270}{\epsfig{figure=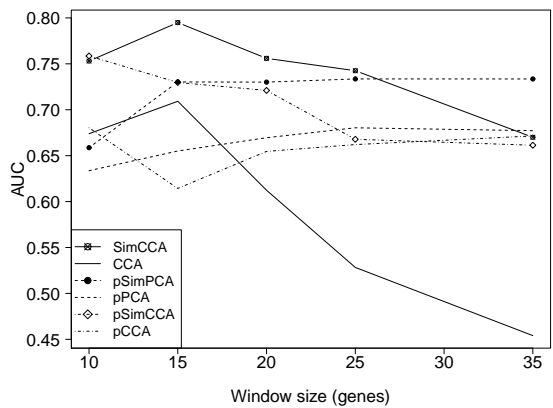,width=6cm}}}
\end{minipage}
\caption{AUC comparison.}
\label{fig:auc}
\end{figure}

\begin{figure}[htb]
\begin{minipage}[b]{1.0\linewidth}
  \centering
 \centerline{\rotatebox{270}{\epsfig{figure=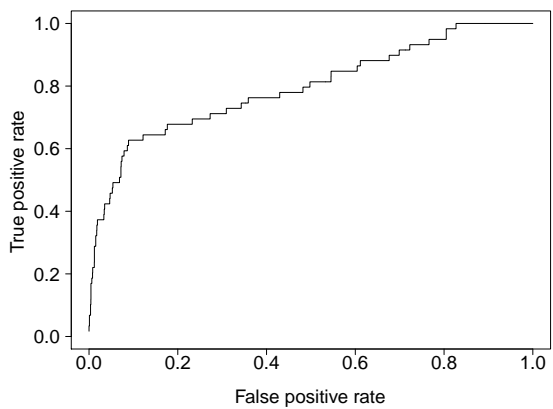,width=6cm}}}
\end{minipage}
\caption{ROC curve for the results from correlation-based SimCCA with
  a 15-gene sliding window.}
\label{fig:roc}
\end{figure}

Results from the correlation-based SimCCA are illustrated for
chromosome arm 17q in Fig.~\ref{fig:17q}, where SimCCA highlights a
known cancer-associated region. The Figure shows the dependency score
for the correlation-based SimCCA with a sliding window of 15 genes
genes along the chromosome arm.  The correlation-based and
probabilistic approaches were \-compared in various window sizes (10,
15, 20, 25, and 35 genes). In each experiment, the gene list ordered
by the dependency measure was compared to an expert-curated list of 59
gastric-cancer associated genes in our investigated data set
\cite{Myllykangas08}. 

The correlation-based and probabilistic models were compared with
respect to their ability to detect the known cancer genes, measured
with the AUC value of the ROC curve for each method.  Results are
summarized in Fig.~\ref{fig:auc}.  The best AUC value (0.79) was
obtained with a chromosomal window of 15 genes for the
correlation-based SimCCA that directly maximizes the correlations
assuming identical projections (Eq.~(\ref{eq:simcca})). The
corresponding ROC curve is shown in Fig.~\ref{fig:roc} and presents
the tradeoff between true and 'false' positive findings along the
ordered gene list. While a large proportion of the most significant
findings are in fact known cancer genes, the remaining findings with
no known associations to gastric cancer are promising candidates for
further studies; among the 100 genes with highest dependencies between
gene expression and copy number in their chromosomal neighborg\-hood,
30\% of the corresponding regions had previously known association
with gastric cancer, while the proportion in the whole data set is
5\%.

The constrained dependency detection methods introduced in this paper
outperformed the unconstrained models in most cases. The improved
detection performance of the constrained models is likely explained by
their ability to reduce overfitting. Interestingly, the most
constrained probabilistic model, pSimPCA, outperforms the other
approaches in the highest-dimensional case. In contrast, the
performance of correlation-based CCA decreases steadily with
increasing dimensionality (window size) as the number of samples
(patients) remains fixed to 51.

In our particular application, gene expression and copy number are
expected to have strong linear correlations in cancer-associated
chromosomal regions.  Correlation-based approach is therefore directly
suited for the cancer gene detection task and it has also fewer
parameters than the probabilistic versions. However, the performance
of correlation-based SimCCA reduces with increasing dimensionality. A
likely explanation is that the correlation-based version models also
some of the data set-specific effects, which is emphasized in
higher-dimensions. The probabilistic formulations provide an
alternative way to bring in prior knowledge of the relationships in a
principled framework. A potential advantage of the probabilistic
approaches is that they have an explicit model for distinguishing the
shared signal from data set-specific variation.

\subsection{Biomedical interpretation of the findings}

The results obtained using the SimCCA algorithm are in general
concordant with the output from signal-to-noise sta\-tistics and
random permutation method that was applied previously to analyze the
same data \cite{Myllykangas08, Hautaniemi04}. The advantage of the
current method is that it combines the signal across adjacent genes
within a particular chromosomal region already in the modeling step.
Probabilistic SimCCA estimates the strongest shared signal between the
data sets and ignores other variation using explicit modeling
assumptions.  Probabilistic versions also provide a measure of the
amplification effect for each patient which allows robust
identification of small patient groups with profound amplification
effects that would be missed in previous permutation-based tests due
to low event frequency.

In concordance with the previous analyses, the chromosomal area
showing the most significant correlation between the gene copy number
and expression was 17q12-q21 (Fig.~\ref{fig:17q}). There are a number
of potential target genes in that region, including {\it ERBB2} and
{\it PPP1R1B}, which show clinical and biological relevance. The {\it
  ERBB2} gene encodes a transmembrane tyrosine kinase receptor, which
is a target of Herceptin. This monoclonal antibody specifically
inactivates the overexpressed {\it ERBB2} protein and is used to treat
metastatic breast cancer patients. The expression of {\it PPP1R1B} has
been shown to be associated with repression of programmed cell death
and increase the survival of the cancer cells in upper
gastrointestinal tract cancers \cite{Belkhiri05}.

Another genomic region with correlated gene copy number and expression
changes is 10q26, and {\it FGFR2} was identified as one of the
putative target genes of that region.  It was recently shown that in a
set of gastric cancer cell lines, {\it FGFR2} amplification is driving
the cell proliferation and promoting cancer cell
survival. Furthermore, inhibition of the {\it FGFR2} protein by small
molecules retained the growth arresting and apoptotically active
phenotype \cite{Kunii08}. The detected 1q22 region harbors the {\it
  MUC1} gene, whose expression was shown to be associated with the
intestinal subtype of gastric cancer \cite{Myllykangas08}. The 20q is
one of the most frequently amplified chromosomal regions in gastric
cancer. However, despite of high frequency of the amplifications the
target genes in that area remain to be described. Our analysis
pinpointed the strongest correlating loci to 20q13.12 and
significantly narrow the list of putative target genes.

Some of the detected chromosomal regions did not have known
association with gastric cancer; we are currently investigating these
results more closely. The current application shows promising
performance in detecting functional copy number changes, but
biomedical studies provide also a number of other potential
applications.  For example, an increasing number of paired data sets
are available in the future for studying the relationships between
methylation, single-nucleotide polymorphisms, miRNAs, and other genomic
features.

\section{Discussion}

We have introduced methods that regularize CCA solutions by taking
into account similarity constraints. The methods assume that the
dependencies between the different views are visible in the same
dimensions, that is, the projection matrices are similar. We
introduced the constraints to standard CCA, resulting in a quick
method that helps in solving the ``small $n$ large $p$ problem'',
where $n$ is the number of samples and $p$ their dimensionality.

If $n$ is very small compared to $p$, even the constrained CCA may not
be sufficient, and we introduced a Bayesian variant into which further
prior knowledge can be easily inserted, and which is capable of
rigorously handling uncertainty in the data. While we only compare
SimCCA and CCA in the present work, the probabilistic formulation
allows smooth tradeoff between these two extremes, which is
potentially useful in many applications.

Importantly, the constrained approaches for dependency detection can
be directly applied in practical tasks in knowledge discovery; good
results were obtained in a promising medical application on
searching for potential cancer genes by detecting dependencies between
gene expression and DNA copy number changes of the genes.

\subsubsection*{Acknowledgements: The project was funded by Tekes
MultiBio project. LL and SK belong to the Adaptive Informatics
Research Centre and Helsinki Institute for Information Technology
HIIT. LL is funded by the Graduate School of Computer Science and
Engineering. SK is partially supported by EU FP7 NoE PASCAL2, ICT
216886.}

\bibliographystyle{plain}

\end{document}